# A direct method for estimating a causal ordering in a linear non-Gaussian acyclic model

**Shohei Shimizu**
ISIR
Osaka University
Japan

**Aapo Hyvärinen**
Dept. Comp. Sci.
Dept. Math. and Stat.
University of Helsinki & HIIT
Finland

**Yoshinobu Kawahara**
Dept. Math. and Comp. Sci.
Tokyo Institute of Technology
Japan

**Takashi Washio**
ISIR
Osaka University
Japan

**Abstract**

Structural equation models and Bayesian networks have been widely used to analyze causal relations between continuous variables. In such frameworks, linear acyclic models are typically used to model the data-generating process of variables. Recently, it was shown that use of non-Gaussianity identifies a causal ordering of variables in a linear acyclic model without using any prior knowledge on the network structure, which is not the case with conventional methods. However, existing estimation methods are based on iterative search algorithms and may not converge to a correct solution in a finite number of steps. In this paper, we propose a new direct method to estimate a causal ordering based on non-Gaussianity. In contrast to the previous methods, our algorithm requires no algorithmic parameters and is guaranteed to converge to the right solution within a small fixed number of steps if the data strictly follows the model.

## 1 Introduction

Many empirical sciences aim to discover and understand causal mechanisms underlying their objective systems such as natural phenomena and human social behavior. An effective way to study causal relationships is to conduct a controlled experiment. However, performing controlled experiments is often ethically impossible or too expensive in many fields including social sciences [1], bioinformatics [2] and neuroinformatics [3]. Thus, it is necessary and important to develop methods for causal inference based on the data that do not come from such controlled experiments.

Structural equation models (SEM) [1] and Bayesian networks (BN) [4, 5] are widely applied to analyze causal relationships in many empirical studies. A linear acyclic model that is a special case of SEM and BN is typically used to analyze causal effects between continuous variables. Estimation of the model commonly uses covariance structure of data only and in most cases cannot identify the full structure, *i.e.*, a causal ordering and connection strengths, of the model with no prior knowledge on the structure [4, 5].

In [6], a non-Gaussian variant of SEM and BN called a linear non-Gaussian acyclic model (LiNGAM) was proposed, and its full structure was shown to be identifiable without pre-specifying a causal order of the variables. This feature is a significant advantage over the conventional methods [4, 5]. A non-Gaussian method to estimate the new model was also developed in [6] and is closely related to independent component analysis (ICA) [7]. In the subsequent studies, the non-Gaussian framework has been extended in various directions for learning wider variety of SEM and BN [8, 9]. In what follows, we refer to the non-Gaussian model as LiNGAM model and the estimation method as LiNGAM algorithm.

Most of major ICA algorithms including [10, 11] are iterative search methods [7]. Therefore, the LiNGAM algorithms based on the ICA algorithms need some additional information including initial guess, step sizes and convergence criteria. However, such algorithmic parameters are hard to optimize in a systematic way. Thus, the ICA-based algorithms often get stuck in local optima and may not converge to a reasonable solution if the initial guess or step size is badly chosen [12].

In this paper, we propose a new direct method to estimate a causal ordering of variables in the LiNGAM model without prior knowledge on the structure. The new method derives a reasonable causal order of variables by successively reducing each independent component from given data in the model, and this process is completed in steps equal to the number of the variables in the model. It is not based on iterative search in the *parameter space* and needs no step size or similar algorithmic parameters. It is *guaranteed* to converge to the right solution within a small fixed number of steps if the data strictly follows the model. These features of the new method enable the derivation of a more accurate causal order of the variables in a disambiguated and direct procedure. Once the causal orders of variables is identified, the connection strengths between the variables are easily estimated using some conventional covariance-based methods such as least squares and maximum likelihood approaches [1].

The paper is structured as follows. First, in Section 2, we briefly review LiNGAM model and the ICA-based LiNGAM algorithm. We then in Section 3 introduce a new direct method. The performance of the new method is examined by experiments on artificial data in Section 4, and an experiment on real-world data in Section 5. Conclusions are given in Section 6.

## 2 Background

### 2.1 A linear non-Gaussian acyclic model

In [6], a non-Gaussian variant of SEM and BN, which is called LiNGAM, was proposed. Assume that observed data are generated from a process represented graphically by a directed acyclic graph, *i.e.*, DAG. Let us represent this DAG by a $p \times p$ adjacency matrix $\mathbf{B}=\{b_{ij}\}$ where every $b_{ij}$ represents the connection strength from a variable $x_j$ to another $x_i$ in the DAG. Moreover, let us denote by $k(i)$ a causal order of variables $x_i$ so that no later variable influences any earlier variable. For example, a variable $x_j$ is not influenced by a variable $x_i$, *i.e.*, $b_{ji}=0$, if $k(j) < k(i)$.[1] Further, assume that the relations between variables are linear. Without loss of generality, each observed variable $x_i$ is assumed to have zero mean. Then we have

$$x_i = \sum_{k(j)<k(i)} b_{ij} x_j + e_i, \qquad (1)$$

where $e_i$ is an external influence. All external influences $e_i$ are continuous random variables having *non-Gaussian* distributions with zero means and non-zero variances, and $e_i$ are independent of each other so that there is no unobserved confounding variables [5].

We rewrite the model (1) in a matrix form as follows:

$$\mathbf{x} = \mathbf{Bx} + \mathbf{e}, \qquad (2)$$

where $\mathbf{x}$ is a $p$-dimensional variable vector, and $\mathbf{B}$ could be permuted by simultaneous equal row and column permutations to be strictly lower triangular due to the acyclicity assumption [1]. Strict lower triangularity is here defined as to have a lower triangular structure with all zeros on the diagonal.

We emphasize that $x_i$ is equal to $e_i$ if it is not influenced by any other observed variable $x_j$ ($j \neq i$) inside the model, *i.e.*, all the $b_{ij}$ ($j \neq i$) are zeros. In such a case, an external influence $e_i$ is *observed* as $x_i$. Such an $x_i$ is called an *exogenous observed* variable.[2] Otherwise, $e_i$ is called an *error*. For example, consider the model defined by

$$\begin{aligned} x_1 &= e_1 \\ x_2 &= 1.5x_1 + e_2 \\ x_3 &= 0.8x_1 - 1.5x_2 + e_3, \end{aligned}$$

where $x_1$ is equal to $e_1$ since it is not influenced by either $x_2$ or $x_3$. Thus, $x_1$ is an exogenous observed variable, and $e_2$ and $e_3$ are errors. Note that there *exists at least one exogenous observed variable* $x_i (=e_i)$ due to the acyclicity and the assumption of no unobserved confounders.

### 2.2 Identifiability of the model

We next explain how the connection strengths of the LiNGAM model (2) can be identified as shown in [6]. Let us first solve Eq. (2) for $\mathbf{x}$. Then we obtain

$$\mathbf{x} = \mathbf{Ae}, \qquad (3)$$

where $\mathbf{A} = (\mathbf{I} - \mathbf{B})^{-1}$ is a mixing matrix whose elements are called mixing coefficients and is lower triangular due to the aforementioned feature of $\mathbf{B}$ and the nature of matrix inversion. Since the components of $\mathbf{e}$ are independent and non-Gaussian, Eq. (3) defines the independent component analysis (ICA) model [7], which is known to be identifiable [13].

ICA essentially can estimate $\mathbf{A}$ (and $\mathbf{W} = \mathbf{A}^{-1} = \mathbf{I} - \mathbf{B}$), but has permutation and scaling indeterminacies. ICA actually gives $\mathbf{W}_{ICA}=\mathbf{PDW}$, where $\mathbf{P}$ is an unknown permutation matrix, and $\mathbf{D}$ is an unknown diagonal scaling matrix. But in LiNGAM, the correct permutation matrix $\mathbf{P}$ can be found [6]: the correct $\mathbf{P}$ is the only one that gives no zeros in the diagonal of $\mathbf{DW}$ since $\mathbf{B}$ should be a matrix that can be permuted to be strictly lower triangular and $\mathbf{W} = \mathbf{I} - \mathbf{B}$. Further, one can find the correct scaling of the independent components by using the unity on the diagonal of $\mathbf{W}=\mathbf{I}-\mathbf{B}$. One only has to divide the rows of

---

[1] Note that $k(j)<k(i)$ does not necessarily imply that $x_j$ influences $x_i$. It only implies that $b_{ji}=0$, and $b_{ij}$ can be either zero or non-zero. The causal ordering $k(i)$ only defines a *partial* order of variables, which is enough to define a DAG.

[2] An exogenous variable is defined as a variable that is not influenced by any other variable inside the model. The definition does not require that it is equal to an external influence. However, in the LiNGAM model (2), an exogenous observed variable is always equal to an external influence due to the assumption of no unobserved confounders.

**DW** by its corresponding diagonal elements to obtain **W**. Finally, one can compute the connection strength matrix $\mathbf{B} = \mathbf{I} - \mathbf{W}$.

### 2.3 Original LiNGAM algorithm

The original LiNGAM algorithm presented in [6] is described as follows:

---

(Original) LiNGAM algorithm

1. Given a $p$-dimensional variable vector **x** and its $p \times n$ data matrix **X**, apply an ICA algorithm (FastICA [11] here) to obtain an estimate of **A**.

2. Find the unique permutation of rows of $\mathbf{W} = \mathbf{A}^{-1}$ which yields a matrix $\widetilde{\mathbf{W}}$ without any zeros on the main diagonal. The permutation is sought which minimizes $\sum_i 1/|\widetilde{\mathbf{W}}_{ii}|$.

3. Divide each row of $\widetilde{\mathbf{W}}$ by its corresponding diagonal element, to yield a new matrix $\widetilde{\mathbf{W}}'$ with all ones on the diagonal.

4. Compute an estimate $\widehat{\mathbf{B}}$ of **B** using $\widehat{\mathbf{B}} = \mathbf{I} - \widetilde{\mathbf{W}}'$.

5. To derive a causal order $k(i)$, find the permutation matrix $\widetilde{\mathbf{P}}$ of $\widehat{\mathbf{B}}$ yielding a matrix $\widetilde{\mathbf{B}} = \widetilde{\mathbf{P}}\widehat{\mathbf{B}}\widetilde{\mathbf{P}}^T$ which is as close as possible to a strictly lower triangular structure. The following approximative algorithm is used, which sets small absolute valued elements in $\widetilde{\mathbf{B}}$ to zero and tests if the resulting matrix is possible to be permuted to be strictly lower triangular:

   (a) Set the $p(p+1)/2$ smallest (in absolute value) elements of $\widehat{\mathbf{B}}$ to zero.
   (b) Repeat
      i. Test if $\widehat{\mathbf{B}}$ can be permuted to be strictly lower triangular. If the answer is yes, stop and return the permuted $\widehat{\mathbf{B}}$, that is, $\widetilde{\mathbf{B}}$.
      ii. Additionally set the next smallest (in absolute value) element of $\widehat{\mathbf{B}}$ to zero.

---

### 2.4 Potential problems of original LiNGAM

The original ICA-based LiNGAM algorithm has several potential problems: i) Most ICA algorithms including FastICA [11] and gradient-based algorithms [10] may not converge to a correct solution in a finite number of steps if the initially guessed state is badly chosen [12] or if the step size is not suitably selected for those gradient-based methods. The appropriate selection of such algorithmic parameters is not easy. In contrast, our algorithm proposed in the next section is guaranteed to converge in a fixed number of steps equal to the number of variables. ii) The permutation algorithms in Steps 2 and 5 are not scale-invariant. Hence they could give a different or *even wrong* ordering of variables depending on scales or standard deviations of variables especially when they have a wide range of scales. However, scales are essentially not relevant to the ordering of variables. Though such bias would vanish for large enough sample sizes, for practical sample sizes, an estimated ordering could be affected when variables are normalized to make unit variance, and hence the derivation of a reasonable ordering becomes quite difficult.

## 3 A direct method: DirectLiNGAM

### 3.1 Identification of an exogenous variable based on non-Gaussianity and independence

In this subsection, we present two lemmas and a corollary that ensure the validity of our algorithm proposed in the next subsection 3.2. The basic idea of our method is as follows. We first find an exogenous variable as the top variable in the causal order by applying Lemma 1. Next, we remove the component of the exogenous variable from the other variables using least squares regression. Then, we show that a LiNGAM model also holds for the residuals (Lemma 2) and that the ordering of the residuals is *equivalent* to that of the corresponding original observed variables (Corollary 1). Therefore, we can find the second top variable in the causal ordering of the original observed variables by analyzing the residuals and their LiNGAM model, *i.e.*, by applying Lemma 1 to the residuals and finding an "exogenous" residual. The repeat of these component removal and causal ordering derives the causal order of the original variables.

**Definition 1 (correlation-faithfulness)** *The distribution of* **x** *is said to be correlation-faithful to the generating graph if correlation and conditional correlation of $x_i$ are entailed by the graph structure, i.e., the zero/non-zero status of $b_{ij}$, but not by special parameter values of $b_{ij}$.*

This concept is motivated by the faithfulness [5]. It is relatively stronger than the original one but would still be acceptable in many cases.

**Lemma 1** *Assume that the input data* **x** *follows the LiNGAM model (2). Further, assume that the distribution of* **x** *is correlation-faithful to the generating graph. Denote by $r_i^{(j)}$ the residuals when $x_i$ are regressed on $x_j$: $r_i^{(j)} = x_i - \frac{\text{cov}(x_i, x_j)}{\text{var}(x_j)} x_j$ ($i \neq j$). Then a variable $x_j$ is exogenous if and only if $x_j$ is independent of its residuals $r_i^{(j)}$ for all $i \neq j$.* □

**Proof** (i) Assume that $x_j$ is exogenous, i.e., $x_j=e_j$. Due to the model assumption and Eq. (3), one can write $x_i=a_{ij}x_j+\bar{e}_i^{(j)}$, $(i\neq j)$, where $\bar{e}_i^{(j)}=\sum_{h\neq i,j}a_{ih}e_h$ and $x_j$ are independent, and $a_{ij}$ is a mixing coefficient from $x_j$ to $x_i$ in Eq. (3). The mixing coefficient $a_{ij}$ is equal to the regression coefficient when $x_i$ is regressed on $x_j$ since $\mathrm{cov}(x_i,x_j)=a_{ij}\mathrm{var}(x_j)$. Thus, the residual $r_i^{(j)}$ is equal to the corresponding error term, i.e., $r_i^{(j)}=\bar{e}_i^{(j)}$. Thus, $x_j$ and $r_i^{(j)}(=\bar{e}_i^{(j)})$ are independent.

(ii) Assume that $x_j$ is not exogenous. Then there always exist such variables $x_h$ that $x_j=\sum_{k(h)<k(j)}b_{jh}x_h+e_j$ $(b_{jh}\neq 0)$, where $x_h$ and $e_j$ are independent. Let $P_j$ denote the set of such parent variables of $x_j$. Then, for a variable $x_i\in P_j$, we have

$$
\begin{aligned}
r_i^{(j)} &= x_i - \frac{\mathrm{cov}(x_i,x_j)}{\mathrm{var}(x_j)}x_j \\
&= \left\{1-\frac{b_{ji}\mathrm{cov}(x_i,x_j)}{\mathrm{var}(x_j)}\right\}x_i - \frac{\mathrm{cov}(x_i,x_j)}{\mathrm{var}(x_j)}e_j \\
&\quad -\frac{\mathrm{cov}(x_i,x_j)}{\mathrm{var}(x_j)}\sum_{x_h\in P_j, h\neq i}b_{jh}x_h. \quad (4)
\end{aligned}
$$

Recall that the correlation-faithfulness is assumed. Since $x_i$ has a non-zero connection strength $b_{ji}$ to $x_j$, their covariance $\mathrm{cov}(x_i,x_j)$ is not zero, and hence the coefficient of $e_j$ is not zero. According to this fact and another fact that $e_j$ is independent of all $x_h, x_i\in P_j$ but dependent on $x_j$, $x_j$ and $r_i^{(j)}$ are dependent.

From (i) and (ii), the lemma is proven. ∎

**Lemma 2** *Assume the assumptions of Lemma 1. Further, assume that a variable $x_j$ is exogenous. Denote by $\mathbf{r}^{(j)}$ a (p-1)-dimensional vector that collects the residuals $r_i^{(j)}$ when all $x_i$ of $\mathbf{x}$ are regressed on $x_j$ $(i\neq j)$. Then a LiNGAM model holds for the residual vector $\mathbf{r}^{(j)}$: $\mathbf{r}^{(j)} = \mathbf{B}^{(j)}\mathbf{r}^{(j)} + \mathbf{e}^{(j)}$, where $\mathbf{B}^{(j)}$ is a matrix that can be permuted to be strictly lower-triangular by a simultaneous row and column permutation, and elements of $\mathbf{e}^{(j)}$ are non-Gaussian and mutually independent.* □

**Proof** Without loss of generality, assume that $\mathbf{B}$ in the LiNGAM model (2) is already permuted to be strictly lower triangular and that $x_j=x_1$. Note that $\mathbf{A}$ in Eq. (3) is also lower triangular (although its diagonal elements are all unities). Since $x_1$ is exogenous, $a_{i1}$ are equal to the regression coefficients when $x_i$ are regressed on $x_1$ $(i\neq 1)$. Therefore, after removing the effects of $x_1$ from $x_i$ by least squares estimation, one gets the first column of $\mathbf{A}$ to be a zero vector, and the residuals $r_i^{(1)}$ are not influenced by $x_1$. (Due to the correlation-faithfulness, the effect of $x_1$ is always removed from $x_i$ when $x_1$ influences $x_i$ because it does not happen that their covariance $\mathrm{cov}(x_i,x_1)$ or the regression coefficient from $x_1$ to $x_i$ is accidentally zero due to a combined effect of multiple pathways.) Thus, we again obtain a lower triangular mixing matrix $\mathbf{A}^{(1)}$ with all unities in the diagonal for the residual vector $\mathbf{r}^{(1)}$ and hence have a LiNGAM model for the vector $\mathbf{r}^{(1)}$. ∎

**Corollary 1** *Assume the assumptions in Lemma 2. Denote by $k_{r^{(j)}}(i)$ the order of $r_i^{(j)}$. Recall that $k(i)$ denotes the order of $x_i$. Then, the ordering of the residuals is equivalent to that of corresponding original observed variables: $k_{r^{(j)}}(l)<k_{r^{(j)}}(m) \Leftrightarrow k(l)<k(m)$.*

**Proof** As shown in the proof of Lemma 2, when the effect of an exogenous variable $x_1$ is removed from the other observed variables, the second to $p$-th columns of $\mathbf{A}$ remain the same, and the submatrix of $\mathbf{A}$ formed by deleting the first row and the first column is still lower triangular. This shows that the ordering of the other variables is not changed and proves the corollary. ∎

Lemma 2 indicates that the LiNGAM model for the $(p-1)$-dimensional residual vector $\mathbf{r}^{(j)}$ can be handled as a new input model, and Lemma 1 can be further applied to the model to derive the next exogenous variable (the next exogenous residual in fact). This process can be repeated until all variables are ordered, and the resulting order of the variable subscripts shows the causal order of the original observed variables according to Corollary 1.

Next, we define an independence (not merely uncorrelatedness) measure.[3] Let us denote by $U$ the set of the subscripts of variables $x_i$, i.e., $U=\{1,\cdots,p\}$. We use the following statistic to evaluate nonlinear correlation between a variable $x_j$ and its residuals $r_i^{(j)} = x_i - \frac{\mathrm{cov}(x_i,x_j)}{\mathrm{var}(x_j)}x_j$ when $x_i$ is regressed on $x_j$:

$$
T(x_j;U) = \sum_{i\in U, i\neq j}\left[|\mathrm{corr}\{g(r_i^{(j)}),x_j\}| + |\mathrm{corr}\{r_i^{(j)},g(x_j)\}|\right], \quad (5)
$$

where $g$ is a nonlinear and non-quadratic function, e.g., $g(\cdot)=tanh(\cdot)$. The statistic $T$ in Eq. (5) is zero if $x_j$ and $r_i^{(j)}$ are independent. Strictly speaking, independence is a much stronger condition than requiring the statistic $T$ to be zero. However, in many cases evaluating such a nonlinear correlation as Eq. (5) would work well enough as implied in the ICA literature [7]. This also leads to a fair comparison of our method with the original LiNGAM algorithm that uses FastICA [11]. FastICA minimizes almost the same type of nonlinear correlation of estimated independent components $E\{g(\hat{e}_i)\hat{e}_j\}$ ($i{\neq}j$) in absolute value sense [15]. More sophisticated nonparametric independence measures [16, 17] have been proposed. We can use them instead of the statistic $T$ in Eq. (5) to evaluate independence in our framework when needed.

---

[3]Least squares regression gives residuals always uncorrelated with but not necessarily independent of explanatory variables [14].

### 3.2 DirectLiNGAM algorithm

We now propose a new direct algorithm called DirectLiNGAM to estimate a causal ordering and the connection strengths in the LiNGAM model (2) under the correlation-faithfulness assumption:

---

DirectLiNGAM algorithm

1. Given a $p$-dimensional variable vector $\mathbf{x}$, a set of its variable subscripts $U$ and a $p \times n$ data matrix of the variable vector as $\mathbf{X}$, initialize an ordered list of variables $K = \emptyset$ and $m := 1$.

2. Repeat until $p-1$ subscripts are appended to $K$:

   (a) Perform least squares regressions of $x_i$ on $x_j$ for all $i \in U - K$ ($i \neq j$) and derive the residual vectors $\mathbf{r}^{(j)}$ and its residual data matrix $\mathbf{R}^{(j)}$ from the data matrix $\mathbf{X}$ for all $j \in U - K$. Find a variable $x_m$ that is most independent of its residuals:
   $$x_m = \arg \min_{j \in U-K} T(x_j; U - K), \qquad (6)$$
   where $T$ is the independence measure defined in Eq. (5).

   (b) Append $m$ to the end of $K$.

   (c) Let $\mathbf{x} := \mathbf{r}^{(j)}$, $\mathbf{X} := \mathbf{R}^{(j)}$ and $m := m + 1$.

3. Append the remaining variable to the end of $K$.

4. Construct a strictly lower triangular matrix $\mathbf{B}$ by following the order in $K$, and estimate the connection strengths $b_{ij}$ by using some conventional covariance-based regression such as least squares and maximum likelihood approaches on the original variable vector $\mathbf{x}$ and the original data matrix $\mathbf{X}$.

---

We note that our DirectLiNGAM can be applied to situations with more variables than observations ($p>n$), whereas the original LiNGAM would fail completely. However, in Step 4, a small tip is necessary to estimate connection strengths when regression analysis is performed to variables that have more parent variables than observations. In that case, lasso-type estimation methods [18] would be useful.

### 3.3 Computational complexity

Here, we consider the computational complexity of DirectLiNGAM compared with the original LiNGAM with respect to sample size $n$ and number of variables $p$. A dominant part of DirectLiNGAM is to compute Eq. (5) for each $x_j$ in Step 2(a). Since it requires $O(np^2)$ operations in $p-1$ iterations, complexity of the step is $O(np^3)$. Another dominant part is the regression to derive the matrix $\mathbf{B}$ in Step 4. The complexity of many representative regressions including the least square algorithm is $O(np^3)$. Hence, we have a total budget of $O(np^3)$. Meanwhile, the original LiNGAM requires $O(p^4)$ time to find a causal order in Step 5. Complexity of an iteration in FastICA procedure at Step 1 is known to be $O(np^2)$. Assuming a constant number $C$ of the iterations in FastICA steps, the complexity of the original LiNGAM is considered to be $O(Cnp^2 + p^4)$. Though general evaluation of the required iteration number $C$ is difficult, it is usually conjectured to grow linearly with regards to $p$. Hence the complexity of the original LiNGAM is presumed to be $O(np^3 + p^4)$. Accordingly, the computational cost of DirectLiNGAM is considered to be almost same with or, especially in cases where $p$ increases, superior to that of the original LiNGAM because of its order $O(p^3)$ against $O(p^4)$. In fact, while the original LiNGAM requires $p \ll n$ to perform FastICA, DirectLiNGAM requires the computation of covariance between variables only and can be carried out even when $p > n$ that sometimes occurs in real-world applications. Moreover, we here emphasize the fact that DirectLiNGAM has guaranteed convergence in a fixed number of steps and of known complexity, whereas for typical ICA algorithms, the run-time complexity and the very convergence are not guaranteed.

## 4 Simulations

We randomly generated 501 datasets under each combination of number of variables $p$ and sample size $n$ ($p$=10, 20, 50, 100; $n$=30, 50, 80, 200, 500, 1000, 2000, 5000) as follows:

1. We randomly constructed a $p \times p$ strictly lower-triangular matrix $\mathbf{B}$ so that standard deviations of variables $x_i$ owing to parent variables ranged in the interval $[0.5, 1.5]$. Either of a fully connected network or a sparse network was randomly created. We also randomly selected standard deviations of the external influences $e_i$ from $[0.5, 1.5]$.

2. We generated data with sample size $n$ by independently drawing the external influence variables $e_i$ from various non-Gaussian distributions with zero mean and unit variance. We first generated

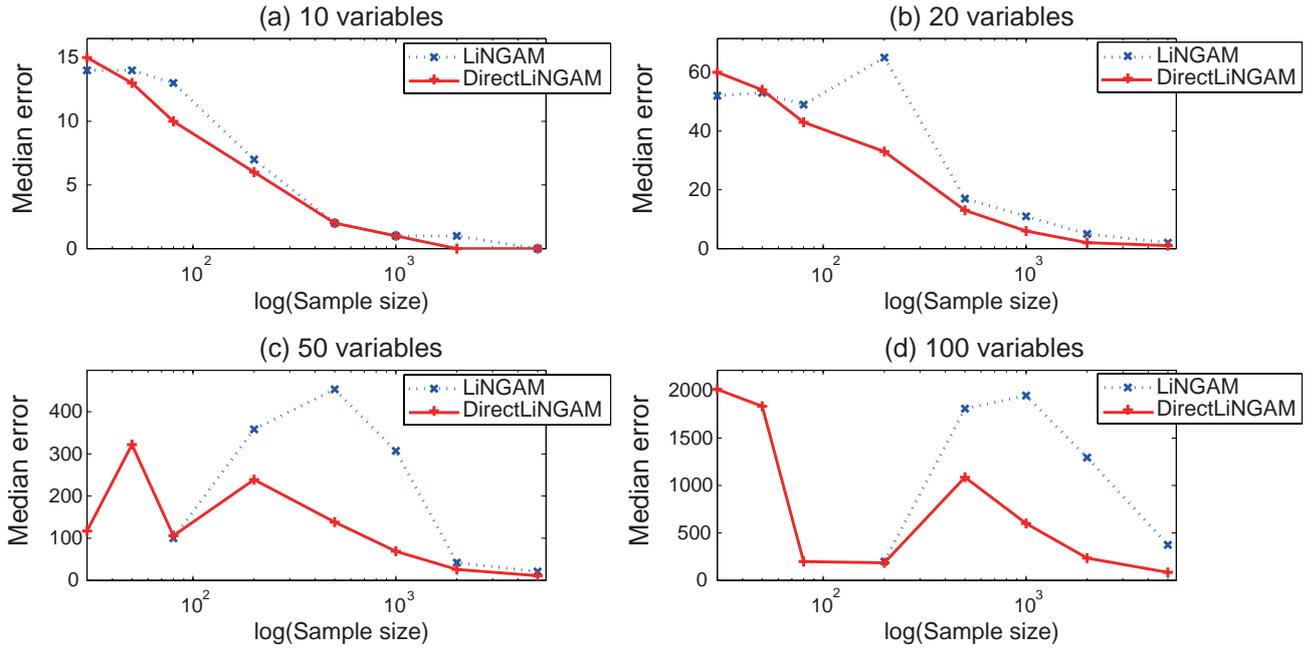

Figure 1: Median numbers of errors in estimated $K$ of original LiNGAM and DirectLiNGAM under (a) 10 variables; (b) 20 variables; (c) 50 variables; (d) 100 variables.

Gaussian variables $z_i$ with zero means and unit variances and subsequently transformed it to non-Gaussian variables by $e_i = \text{sign}(z_i)|z_i|^{q_i}$. The nonlinear exponents $q_i$ were randomly selected from the interval $[0.5, 0.8] \cup [1.2, 2.0]$. Nonlinear exponents $q_i$ selected from $[0.5, 0.8]$ gave sub-Gaussian variables, and exponents selected from $[1.2, 2.0]$ provided super-Gaussian variables. Finally, the transformed variables were standardized to have zero means and unit variances.

3. The values of the observed variables $x_i$ were generated according to the LiNGAM model (2). Finally, we randomly permuted the order of $x_i$.

This is the *same* procedure as the one used to test the original LiNGAM in [6], and we did *not* do anything to make parameter values satisfy the correlation-faithfulness assumption.

Then we tested the original LiNGAM and our DirectLiNGAM on the datasets using the same non-linearity $g(\cdot) = tanh(\cdot)$ for the statistic $T$ in Eq. (5) and FastICA [11] respectively to evaluate nonlinear correlation. For each trial, we first permuted the *true* connection strength matrix $\mathbf{B}$ according to estimated orderings $K$ by the original LiNGAM and our DirectLiNGAM. We then counted the number of errors, *i.e.*, how many elements in its strictly upper triangular part are not zeros. If the ordering is correctly estimated, the elements in the strictly upper triangular part are *all zeros*. The medians of the numbers of non-zero strictly upper triangular elements were plotted in Fig. 1. The median errors of our DirectLiNGAM were much smaller than the original LiNGAM for most of the practical experimental conditions, although the median errors of both methods were rather large for smaller sample sizes.

In Fig. 2, numbers of non-zero strictly upper triangular elements in the permuted true $\mathbf{B}$ based on estimated $K$ under 50 variables and 500 observations are shown in form of boxplots.[4] We also computed the distance between the true $\mathbf{B}$ and ones estimated by the original LiNGAM and DirectLiNGAM using the Frobenius norm defined as $\sqrt{\text{trace}\{(\mathbf{B}_{true} - \widehat{\mathbf{B}})^T(\mathbf{B}_{true} - \widehat{\mathbf{B}})\}}$. The matrices $\mathbf{B}$ were estimated by using FastICA in the original LiNGAM and the least square regression in DirectLiNGAM. Boxplots of the distances are also shown in Fig. 3. DirectLiNGAM was better in both median numbers of errors in $K$ and distances of $\mathbf{B}$ and has their smaller variability than the original LiNGAM.

There was a substantial variability in the medians for higher variable dimensions. It is probably because those sample sizes would not be large enough for the corresponding variable dimensions, and many of exper-imental conditions including network structures and distributions of external influences were randomized. Future work would conduct more extensive simulations to study the performance of our method for different types of non-Gaussian external influences and to compare computational costs with the original LiNGAM.

---

[4]In a boxplot, a red line indicates a median $q_1$, the bottom edge of a blue box a 25th percentile $q_2$, its top edge a 75th percentile $q_3$, a lower black line a percentile $q_2 - 1.5(q_3 - q_2)$ or the minimum and an upper black line a percentile $q_3 + 1.5(q_3 - q_2)$ or the maximum. The height of the blue box measures dispersion of data points.

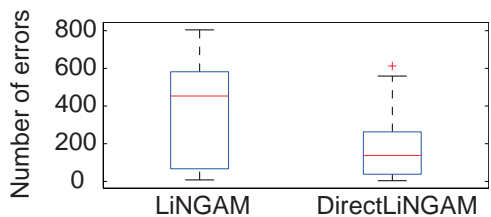

Figure 2: Boxplots of numbers of non-zero strictly upper triangular elements based on estimated $K$ of original LiNGAM and DirectLiNGAM under 50 variables and 500 observations.

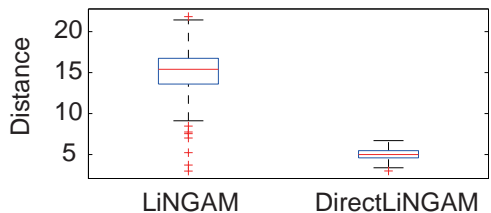

Figure 3: Boxplots of distances between true **B** and estimated **B** of original LiNGAM and DirectLiNGAM under 50 variables and 500 observations.

Our method repeats identification of an exogenous variable and its removal to derive a causal ordering. Early mistakes in the repeat might make more errors. Such mistakes are likely when an endogenous variable receives a quite small influence(s) close to zero from its parent variable(s), or the model is close to being not correlation-faithful. An important question for future research is to investigate how sensitive to early mistakes our method is and, eventually, how it can be alleviated.

## 5 Application to magnetoencephalography data

As an illustration of the applicability of the method on real data, we applied it on magnetoencephalography (MEG) data, *i.e.*, measurements of the electric activity in the brain. The raw data consisted of the 306 MEG channels measured by the Vectorview helmet-shaped neuromagnetometer (Neuromag Ltd., Helsinki, Finland) in a magnetically shielded room at the Brain Research Unit, Low Temperature Laboratory, Helsinki University of Technology. The measurements consisted of 300 seconds of resting state brain activity earlier used in [19]. The subject was sitting with eyes closed, and did not perform any specific task nor was there any specific sensory stimulation.

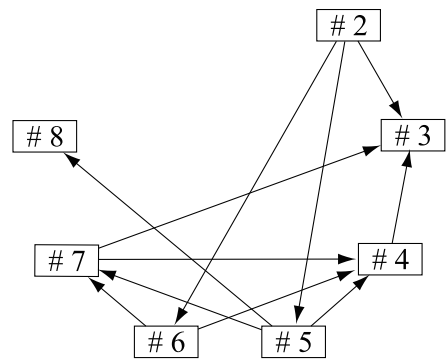

Figure 4: The estimated network by original LiNGAM.

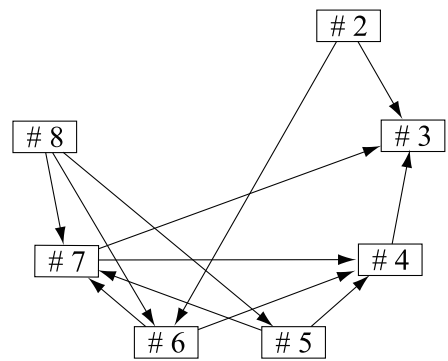

Figure 5: The estimated network by DirectLiNGAM.

First, we performed a blind separation of sources using the method called Fourier-ICA [19]. This gave the nine sources obtained in [19]. For further analysis we chose those sources that were clearly localized in the brain, which meant discarding two of the estimated sources #1 and #9. Our goal was to analyze the causal relations between the powers of the source, so we divided the data into windows of length of one second (half overlapping, *i.e.*, the initial points were at a distance of 0.5 seconds each) and computed the local standard deviation of every source in each window. This gave a total of 604 observations of a seven-dimensional random vector, on which we applied our method.

For each of the orderings of variables estimated by the original LiNGAM and DirectLiNGAM, we estimated the connection strengths and computed their 99% confidence intervals by using least squares regression and bootstrapping [20]. The estimated networks by the original LiNGAM and DirectLiNGAM were shown in Figures 4 and 5 respectively, where only significant arrows were shown with 1% significance level. The difference between the two estimated networks is essentially only that the former indicates the endogeneity of

#8 while the latter shows its exogeneity. Variable #8 is probably exogenous since DirectLiNGAM considers that it is exogenous, and DirectLiNGAM is specifically adapted, by construction, to detect exogenous variables. Original LiNGAM does not consider variable #8 exogenous, possibly because it got stuck in a local optima. Variable #8 is, in fact, an interesting and complicated source because its frequency contents are different from all the other sources.

## 6 Conclusion

We presented a new estimation algorithm for the LiNGAM model (2) that has guaranteed convergence in a fixed number of steps and known computational complexity unlike most ICA methods. This is the first algorithm specialized to estimate the LiNGAM model. Simulations implied that the new method often provides much better statistical performance than state of the art methods based on ICA. In a real-world application to MEG data, a promising result was also obtained. A drawback would be the correlation-faithfulness assumption, but the simulations implied that it might be not very problematic in practice.

**Acknowledgements**

We are very grateful to Pavan Ramkumar and Lauri Parkkonen for providing the MEG data, Hidetoshi Shimodaira and Yusuke Komatsu for interesting discussion and five anonymous reviewers for helpful comments. S.S. was partially supported by MEXT Grant-in-Aid for Young Scientists #21700302. A.H. was partially supported by the Academy of Finland Centre of Excellence for Algorithmic Data Analysis. Y.K. was partially supported by JSPS Grant-in-Aid for Young Scientists #20800019. T.W. was partially supported by Grant-in-Aid for Scientific Research (A) #19200013. S.S. and Y.K. were partially supported by JSPS Global COE program 'Computationism as a Foundation for the Sciences'. This work was partially carried out at Department of Mathematical and Computing Sciences and Department of Computer Science, Tokyo Institute of Technology.